\documentclass[runningheads]{llncs}

\usepackage{hyperref}
\usepackage{caption}
\usepackage{graphicx}
\usepackage{comment}
\usepackage{multirow}
\usepackage{enumitem}
 
\begin{document}
\title{Unsupervised routine discovery in egocentric photo-streams}


\author{Estefania Talavera\inst{1,2} \and Nicolai Petkov\inst{1} \and Petia Radeva\inst{2}}
%
%
\institute{Bernoulli Institute for Mathematics, Computer Science and Artificial Intelligence, University of Groningen, The Netherlands \and
Department of Mathematcis and Computer Science, University of Barcelona, Spain}
\maketitle              
%

\begin{abstract}
The routine of a person is defined by the occurrence of activities throughout different days, and can directly affect the person's health. In this work, we address the recognition of routine related days. To do so, we rely on egocentric images, which are recorded by a wearable camera and allow to monitor the life of the user from a first-person view perspective. We propose an unsupervised model that identifies routine related days, following an outlier detection approach. We test the proposed framework over a total of 72 days in the form of photo-streams covering around 2 weeks of the life of 5 different camera wearers. Our model achieves an average of 76\% Accuracy and 68\% Weighted F-Score for all the users. Thus, we show that our framework is able to recognise routine related days and opens the door to the understanding of the behaviour of people.
\end{abstract}

\section{Introduction}

Health professionals are continuously working not only to cure but also prevent diseases of people. Looking for an answer on how the life of people can be improved, promoting good routine lifestyles, a natural question is how can the automatic analysis of people's routines help to improve their lives? \cite{bar2001emotional,gardner2015review}. Routine discovery is a challenging task due to the wide range of combinations of image sequences that can describe our days. Describing routine goes hand in hand with mentioning performed activities by people. For instance, routine related days for a person whose job is to teach will be represented by `commute to work', `teach' or `talk in front of an audience', `meetings', `eating', or similar. In contrast, the day of a policeman patrolling during the day could be described by `driving', `walking outside', `talking to people', `eating', `meeting', among others. Some of the activities that describe a routine related day will be shared by most people, such as `eating', or `meeting', or `talking to people'. However, many of them will depend on their job or responsibilities. Routine differs per person, and it has proven to be difficult to define a standard routine pattern for all people. The automatic classification of routine related scenes can represent a valuable tool for many stakeholders, such as psychologists, who would be able to automatically understand and monitor the behaviour of their patients or clients. This automatic tool will allow them to infer how the detected routines affect the life of people and to develop personalized strategies for a change of behaviour.  
\vspace{-1em}
\begin{figure}[ht!]
\centering
\includegraphics[width=0.8\textwidth]{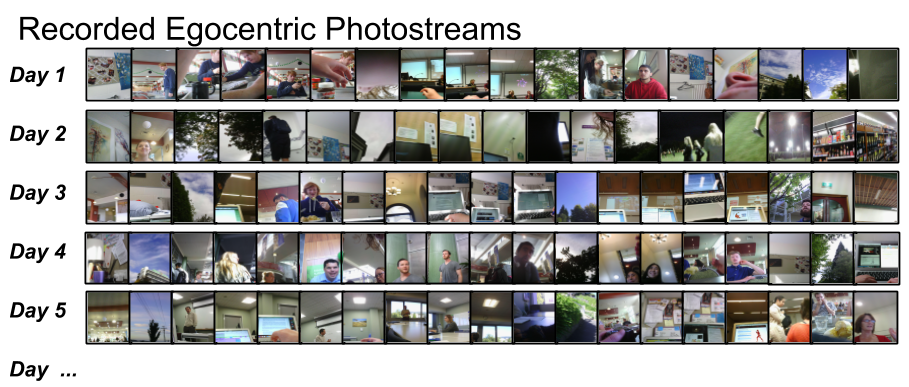}
\caption{Stream of images recorded by one of the camera wearers.}
\label{fig:exampleroutine}
\end{figure}
\vspace{-1em}

Our aim is to recognize routine related days where a person performs common activities of his or her daily living. If we can help people to measure their daily routine, they can try to make changes to their daily habits and the long-term consequences they have on their life. By routine related days, we refer to the days that represent the majority of the days of a person. In consequence, non-routine related days will be described by the basic activities of a daily basis and some other novel activities that are not commonly performed. In this work, we propose a personalized and automatic tool for the discovery of routine related days within recorded photo-streams by a camera wearer. We hypothesize that discovering routine related days can be addressed as a clustering problem where methods such as $k$-means with, for instance, $k=2$ could potentially classify the days in terms of the behaviour they represent.

However, some days present abnormal behaviour. These days correspond to non-routine related days. Most of the time they are not related to each other, which can be interpreted as outliers within the user's recorded photo-streams. Experience has shown that it is difficult to describe what non-routine related days are for a given photo-stream collection. In the context of outlier detection, samples considered as outliers do not form the cluster with higher density when representing the days in a feature space. We propose an unsupervised classification method that assumes that outliers are situated in low-density areas. Outlier detection methods are commonly used in data mining to indicate variability in measurements, errors or novel samples \cite{ding2013anomaly,hodge2004survey}. Among their applications are fraud detection \cite{ghosh1994credit} and satellite image analysis \cite{alvera2012outlier}. However, up to our knowledge for first-time routine detection is defined through an outlier detection approach. Within the available outlier detection algorithms, we propose Isolation Forest algorithm \cite{liu2008isolation}. This method has shown a good performance when detection outliers in multidimensional space, not seeking normal data points but identifying anomalies. Our model is unsupervised because routine differs per person and our aim is to propose a generic model able to discover routine of unknown users. However, since we have the labels of the recorded photo-streams that compose our dataset, we use them to validate if we are able to discover their routine related days.

The closest approaches in computer vision to our aim focus either on scene or activity recognition. However, we cannot characterize a given day as routine related by solely classifying single images from the photo-stream. We observed that information about the patterns of behaviour of a person can be obtained by keeping track of his or her \textit{Activities of Daily Living} (ADLs). These activity patterns can help us expose what is common in the life of the camera wearer. Routine can be inferred when many sample days are recorded since it describes what is common. When aiming to determine a perfect classifier for routine related days, we would need an infinite amount of day samples. Therefore, we think that there is a need for defining a methodology that analyses a collection of images as a group, allowing to compare with other days. We then propose to look for correlation and occurrence of activities throughout the day, and the set of recorded days. 

Hence, the contributions of the paper are three-fold:
\vspace{-0.5em}
\begin{itemize}
\item We address for the first time the problem of routine extraction from egocentric data.
\item We propose an unsupervised and automatic model for the analysis of a routine within recorded egocentric photo-steams. This model is based on the aggregation of the descriptors of the images within the photo-stream.
\item We test our proposed model over a home-made collected egocentric dataset. This dataset describes the daily life of the camera wearers. It is composed of a total of 73000 images, from 72 recorded days by 5 different users. We call it EgoRoutine dataset. 
\end{itemize}

The rest of the paper is organized as follows: in Section \ref{section:PreviousWorks}, we highlight some relevant works related to this topic. Then, in Section \ref{section:proposedmodel} we describe the approach proposed for routine discovery. In Section \ref{section:experimentalsetup}, we describe our newly created EgoRoutine dataset and outline the experiments performed and the obtained results. Finally, in Section \ref{section:conclusions}, we present our conclusions.


\section{Related works}
\label{section:PreviousWorks}

People's routines have been studied in several approaches aiming to characterize patterns of behaviour. Before the emergence of static and wearable sensors, people's daily habits were manually recorded. For instance, ADLs were manually annotated by either individual users and/or specialists, as in \cite{Andersen2004,Habits_life}. In \cite{Andersen2004}, manually recorded information about the ability of someone's ADLs performance was examined, with the aim of classifying the patients' dependence, as either dependent or independent. From another perspective, in \cite{Habits_life}, the authors studied diaries from 70 undergraduate students, who rated the assiduity of activity during the previous month through a questionnaire. The authors highlighted the high standard deviation of the number of behaviours per person. Thus, we conclude that behavioural patterns can be characterized per individual. 

Besides activity recognition, scene recognition has been widely studied from conventional images. In \cite{Kim2010HumanDiscovery} some of the challenges that the recognition of scenes presents were commented; recognizing concurrent activities (same activities performed at the same time), recognizing activities that are shortly interrupted, interpretation when classifying the activity and multiple residents per environment, among others. Nowadays, the incremental use of Convolutional Neural Networks (CNNs) for learning high-level features has shown huge progress in object recognition tasks, mainly due to the availability of large datasets like ImageNet \cite{DengImageNetDatabase}. However, the performance at the scene recognition level is still a challenging task due to the huge range of environments surrounding us and how diverse they can appear. 

\section{Routine Discovery}
\label{section:proposedmodel}

\begin{figure*}[h!]
\centering
\includegraphics[width=\textwidth]{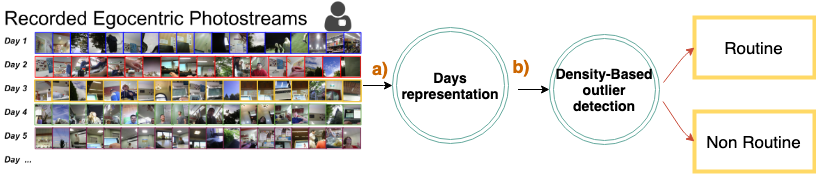}
\caption{The pipeline of the proposed model. Given a set of recorded days, a) they are translated to a set of global or semantic features. Later, b) days are considered as routine or non-routine based on their resemblance.}
\label{fig:pipeline}
\end{figure*}

In this section, we propose an innovative and unsupervised routine discovery method. Its application scheme is given in Fig. \ref{fig:pipeline}. 

\vspace{-1em}
\vspace{-1em}
Our proposed method is based on an outlier detection algorithm. For outlier detection models, an outlier sample is known as a sample outside the 'boundary' of the known classes. In our case, these samples relate to non-routine related days. Hence, we assume that routine related days define a class, of which the samples are close to each other within the feature space. The proposed model indicates routine of the person by detecting the sample days that can be clustered together. In the following subsections, we describe the steps in the proposed pipeline as shown in Fig. \ref{fig:pipeline}.

\subsubsection{a) From days to feature vectors}

As mentioned above, a day is described by a collection of images and takes the form of photo-stream. We address the day classification by translating the recorded photo-streams into feature vectors for their later analysis and comparison. 

Based on the high accuracy recently achieved for the classification of daily activities in egocentric images in \cite{cartas2018batch}, we use their proposed network for the characterization of the recorded days. Given an image, this network classifies it into 21 Activities of Daily Living. A day of the user is represented by $Day = \frac{\sum_{i}^{N} image_{i} }{N}$, where $N$ is the number of images within a day, and $image$ represents the feature vector of the recorded images.

We consider the following descriptors obtained from the collected photo-streams:
\begin{enumerate}
    \item Activity occurrence within the day:  We consider the occurrence of activities throughout the day for the characterization of routine, i.e. bag-of-activities. This feature vector gives an overview of the activities the user performs in a day. However, it does not include temporal information. 
    \item Global descriptors: We use the ResNet CNN model \cite{He2015} to extract global descriptors from the images. We use the activation over the entire image given by the last fully connected layer. Given an image, we obtain a 2048 features vector. 
    \item We concatenate the mentioned features in 1) and 2). 
\end{enumerate}

\subsubsection{b) routine related days recognition}

More specifically, we rely on the unsupervised outlier detection \textit{Isolation Forest} \cite{liu2008isolation} algorithm, and use its available implementation in Scikit-learn \cite{scikit-learn}. It is a tree ensemble method that analyses the density of the space to `isolate' outliers. The algorithm works as follows: 

First, it randomly selects a feature. Then, for the selected feature, it randomly selects a split value between its maximum and minimum value. By recursive partitioning, it can be represented by a tree structure. As the number of trees increases, the algorithm reaches the convergence. The length of the path from the root to the end node can be considered as the number of splittings needed to isolate a sample. By randomly partitioning the data, the paths for anomalies become shorter. Therefore, samples with shorter path lengths are likely to be anomalies. Later, the anomaly score is calculated per sample based on the averaged and normalized distance of the path. Finally, samples considered as outliers have an anomaly score of 1, while samples with values close to 0 are considered as regular.

The \textit{Isolation Forest} algorithm, given a set of \textit{n} samples and an observation \textit{x}, computes the anomaly score \textit{s(x)} as follows:
\begin{equation}
    s(x,n) = 2^{\frac{-E(h(x))}{c(n)}},
\end{equation}   
where h(x) is the path length of a point (x) measured by the number of edges that the point traverses from the root node until the last external node. E(h(x)) corresponds to the  average of h(x) from a collection of isolation trees. c(n) is the average path length, and it is defined as follows:
\begin{equation}
c(n) = 2H(n-1) - (2(n-1)/n),
\end{equation}
where H(i) is the harmonic number and it can be estimated by ln(i) + 0.5772156649 (Euler’s constant).

To \textit{summarize}, given a collection of photo-streams recorded by a camera wearer, our proposed personalized and automatic tool will detect the non-routine related days by computing the density within the feature space. The proposed \textit{Isolation Forest} algorithm considers as routine related days if their samples are in a dense region of samples. In contrast, samples that represent non-routine related days correspond to points in a low-density area. This will have as an output the distinction among days, giving insight into the daily habits and lifestyle of the person.

\section{Experiments}
\label{section:experimentalsetup}

In this section, we describe the experimental setup, the metrics used to evaluate the analysis, and the obtained results.

\subsection{Dataset}

\begin{figure}[h!]
\centering
\includegraphics[width=0.8\textwidth]{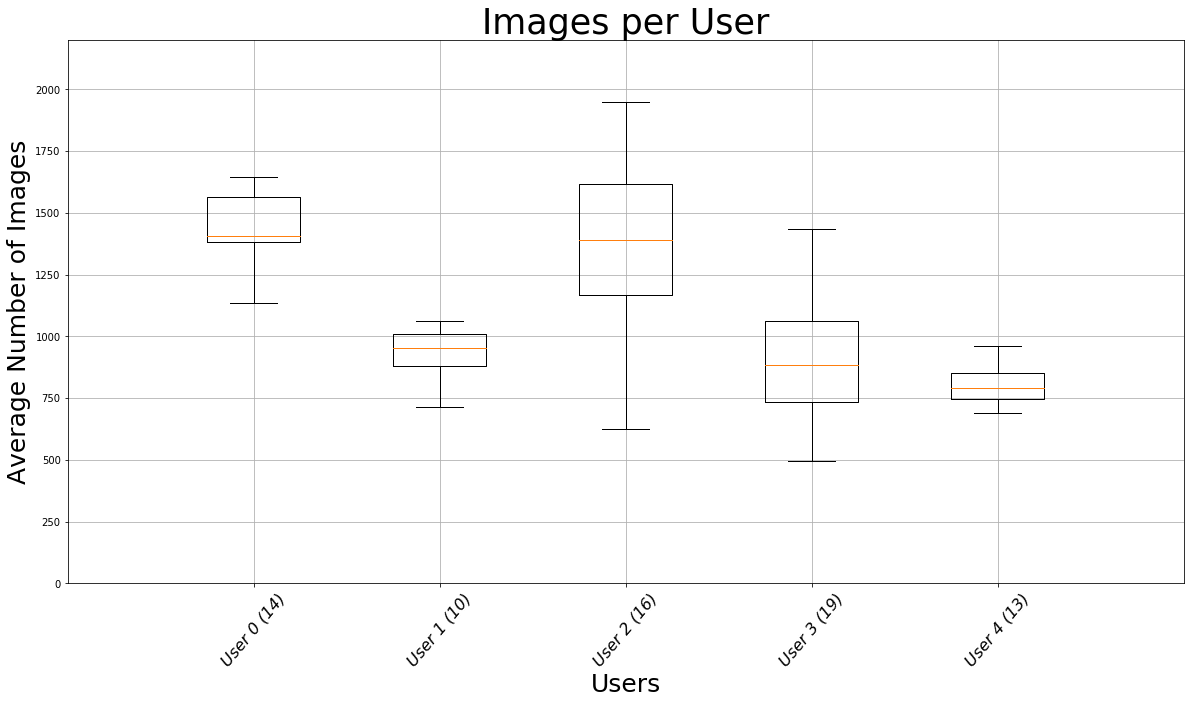}
\caption{Average number of images per recorded egocentric photo-stream. We give the number of collected days per user between parenthesis.}
\label{fig:boxplotdata}
\end{figure}

We collected data from 5 different subjects who were asked to record their daily life during at least a week. To this end, the users worn the \textit{Narrative Clip} camera\footnote{http://getnarrative.com/} fixed to their chest, with a resolution of 2 fpm. The introduced dataset consists of 100k images, from a total of 72 recorded days, see Table \ref{tab:collecteddata_7users}. They captured information about their daily routine, such as the people with whom they interacted, the activities they performed or how often they walked outside. Since there is no training involved in this approach, the whole dataset is analysed by our proposed model.  Moreover, in order to show the variance among collected days, Fig. \ref{fig:boxplotdata} shows the average number of images per day. We can observe how the amount of images differs per day and user.

\begin{table}[t!]
\centering
\resizebox{0.5\textwidth}{!}{
\begin{tabular}{l|ccccc|c}
User ID & \#1 & \#2 & \#3 & \#4 & \#5 & Total \\ \hline
Num Days & 14 & 10 & 16 & 19 & 13 & 72 \\
Images per day & 20k & 8k & 21k  & 13k & 11k & 73k \\ \hline
\end{tabular}}
\caption{Description of the collected Egoroutine dataset by 5 users.}
\label{tab:collecteddata_7users}
\end{table}
\vspace{-1em}

\subsubsection{Process of creating the Ground-truth}

The annotators got the following definition of “\textit{Life routine;} a sequence of actions which are followed regularly, or at specific intervals of time, daily or weekly”. Next, they were shown mosaics of images representing days of the user. They were asked to first have a look at all the mosaics to get an impression of how routine looks like for that specific user. Later, they gave a binary label: routine or non-routine related. 

In Table \ref{GT-labels}, we present the summary of the labels given by the different annotators. From the labelling results, we can deduce that defining what is routine and non-routine is not an easy task. Routine can be easily described in general terms, but it becomes challenging when sequences of images describing a long time period are classified. We can observe how in the majority of cases, the annotators agreed when it comes to label days as routine. However, the non-routine related days are more difficult to perceive leading to disagreement among the annotators. Finally, we have considered as routine related days when $>$4 of the labels agreed. In case of a draw, the day is labelled as non-routine related. Therefore, from a total of 72 recorded days, 51 days are routine related, and 21 are non-routine related. If we extrapolate to a common life scenario, 72 days correspond to almost 15 recorded weeks. If the users followed what could be considered common routine (a week has 5 working days and 2 weekend days or holiday), in 10 weeks we have 20 weekend days and 50 working days. 

\begin{table*}[t!]
\centering
\begin{tabular}{l|c|c|c|c|c}
Class & Six Agree& Five Agree & \begin{tabular}[c]{@{}l@{}}At Least\\ Four Agree\end{tabular} & \begin{tabular}[c]{@{}l@{}}At Least\\ Three Agree\end{tabular} & Total\\ \hline  \hline
All & 34 &	21	 & 11	 & 6 & 72\\ \hline \hline
routine & 28 & 	16  & 7 & 0 & 51 \\
non-routine & 6 & 5 &	4 &	6 & 21\\ \hline 
\end{tabular}
\caption{Summary of the labelling results for the Egoroutine dataset.}
\label{GT-labels}
\vspace{-1em}
\end{table*}
\vspace{-1em}
\subsection{Validation}

We evaluate the performance of the proposed model and compare it with the baseline models by computing the \textit{Accuracy} (Acc), \textit{Recall} (R), \textit{Precision} (P), and \textit{F-Score} metrics, where: $F-Score = 2 \cdot \left( \frac{P \cdot R}{P + R} \right)$  
\textit{Precision} computes the ratio between True Positive (TP) samples and False Positive (FP) samples following: $ TP / (TP + FP) $. \textit{Recall} evaluates the ratio of \textit{TP} and False Negative (FN), showing the ability of the model to find the positive samples, the formula is $ TP / (TP + FN)$. 
Due to the unbalanced dataset we calculate and compare their `macro' and `weighted' mean. The `weighted' mean evaluates the true classification per label, while `macro' calculates the unweighted mean per label. The weighted measures provide the strength of the classifier when applied to unbalanced data. 

\subsection{Experimental setup}

To the best of our knowledge, no previous works have addressed the recognition of routine discovery from egocentric photo-streams. Therefore, we evaluate the performance of the proposed model and compare it with what we introduced as baseline methods. We select several outlier detection algorithms 
namely: Robust Covariance, and One-class SVM. Moreover, we propose to apply unsupervised clustering techniques that allow the identification of outliers or isolation of samples outside the high-density space. These methods allow the recognition of non-similar samples or with non-convex boundaries within the sample collection. Specifically, we evaluate the performance of DBSCAN and Spectral clustering.

Here we give a brief explanation of how these baseline methods work:
\begin{itemize}
    \item Robust Covariance\cite{rousseeuw1999fast}, also called elliptic envelope, assumes that the data follow Gaussian distribution and learns an ellipse. Its drawback is that it degrades when the data is not uni-modal. 
    
    \item One-class SVM \cite{platt1999probabilistic} is an unsupervised algorithm that estimates the support of the dimensional distribution. 

    \item DBSCAN \cite{ester1996density}, short for Density-Based Spatial Clustering of Applications with Noise, finds samples with high density and defines them as the centre of a cluster. From the center, it expands the cluster. Its \textit{eps} parameter determines the maximum distance between samples to be considered as in the same cluster. Outliers are samples that lie alone in low-density regions. 

    \item Spectral Clustering \cite{stella2003multiclass} works on the similarity graph between samples. It computes the first $k$ eigenvectors of its Laplacian matrix and defines a feature vector per sample. Later, $k$-Means is applied to these feature vectors to separate them into $k$ classes. In our case, we set $k=2$, so we evaluate its performance when addressing routine vs non-routine classification.
    
\end{itemize}

For the last two proposed unsupervised model, DBSCAN and Spectral Clustering, the closeness among the recorded days is computed based on their shared similarities, following an all-vs-all strategy. To do so, we use the well-known Euclidean metric. The computed similarity matrix is fed to the unsupervised classifier algorithm for the detection of outliers within the set samples. The outlier detection methods are fed with the feature matrix describing the samples. 

\begin{figure}[ht!]
    \centering 
    \includegraphics[width=0.7\textwidth]{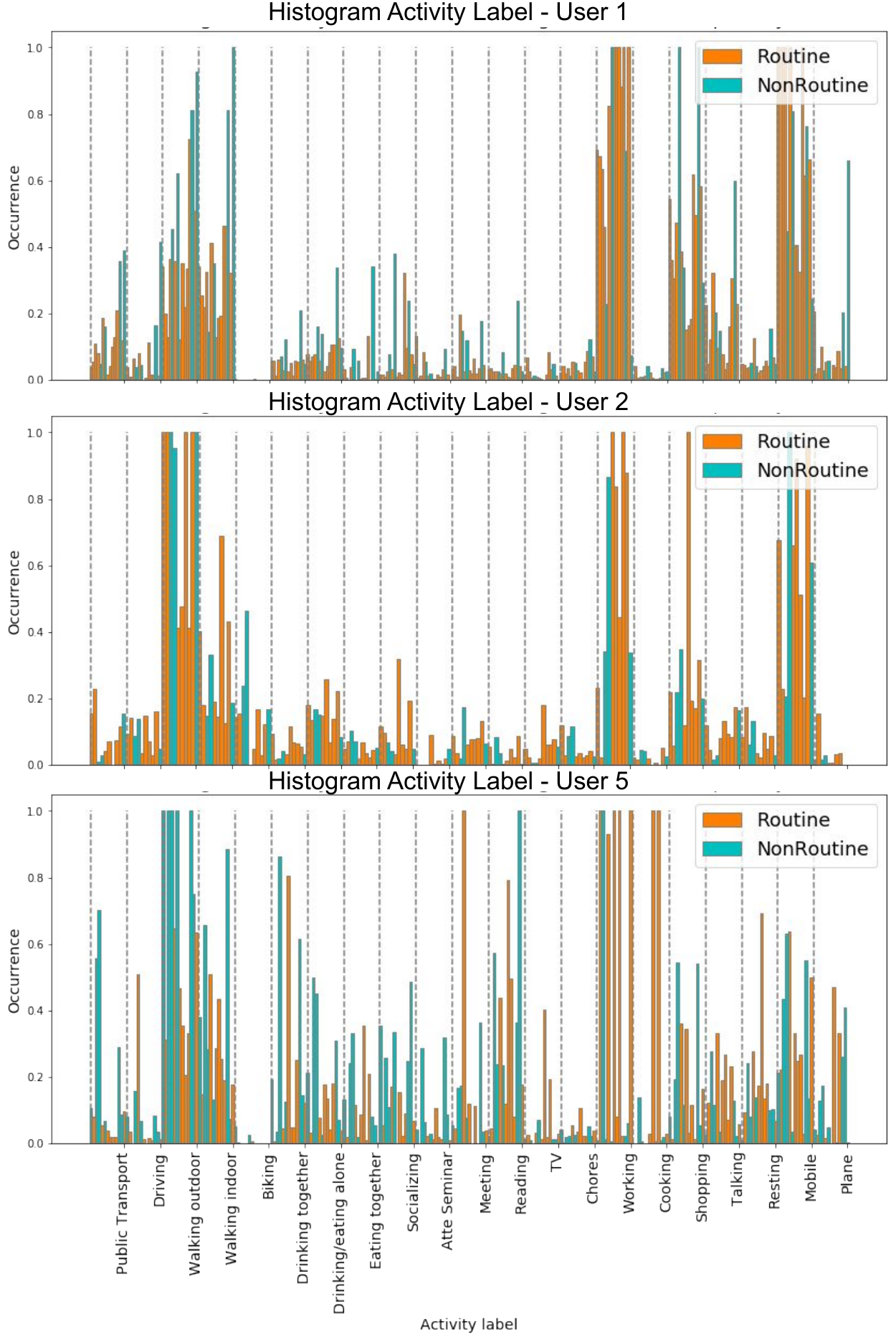}
    \caption{Histograms showing the occurrence of activities throughout the days of 3 of the 5 users that worn the camera. As we can appreciate, some activities are more related to non-routine related days, while `working' and `walking indoor' characterizes routine related days. }
    \label{fig:hist_act_Users}
    \vspace{-1em}
\end{figure}
\vspace{-1em}
\subsection{Results}

We present the obtained classification accuracy at day level for the performed experiments in Table \ref{table:methodsPerformance}. The proposed model, based on the Isolation Forest algorithm and with global features as descriptors of the recorded days, achieved the best performance with respect to the rest of the tested baseline methods. Our model achieves an average of 76\% Accuracy and 68\% Weighted F-Score for all the users, outperforming the rest of the tested methods. The highest performance is when analysing global features, which cover most of the possible present activities.

\begin{table*}[t!]
\begin{center}
\resizebox{0.8\textwidth}{!}{%
\begin{tabular}{l|c|ccccccc}
\multirow{3}{*}{ Methods } & \multirow{3}{*}{ Feature Vector } & \multicolumn{7}{c}{ All Users } \\
& & \multirow{2}{*}{ Acc } & \multicolumn{3}{c}{ Weighted } & \multicolumn{3}{c}{ Macro } \\
& & & F-Score & P & R & F-Score & P & R \\ \hline 
\multirow{3}{*}{ Robust covariance } & Activity Occurrence (Act) & 0.61 & 0.49 & 0.50 & 0.50 & 0.59 & 0.59 & 0.61 \\
& Global Features (Glo) & 0.71 & 0.60 & 0.63 & 0.60 & 0.69 & 0.70 & 0.71 \\
& Act - Glo & 0.54 & 0.39 & 0.39 & 0.41 & 0.52 & 0.51 & 0.54 \\  \hline 
\multirow{3}{*}{ One-Class SVM } & Activity Occurrence (Act) & 0.72 & 0.65 & 0.69 & 0.65 & 0.70 & 0.70 & 0.72 \\
& Global Features (Glo) & 0.67 & 0.56 & 0.60 & 0.57 & 0.64 & 0.67 & 0.67 \\
& Act - Glo & 0.65 & 0.58 & 0.59 & 0.58 & 0.64 & 0.64 & 0.65 \\  \hline 
\multirow{3}{*}{ DBSCAN } & Activity Occurrence (Act) & 0.61 & 0.51 & 0.55 & 0.55 & 0.57 & 0.60 & 0.61 \\
& Global Features (Glo) & 0.69 & 0.41 & 0.34 & 0.50 & 0.56 & 0.48 & 0.69 \\
& Act - Glo & 0.63 & 0.56 & 0.57 & 0.60 & 0.60 & 0.62 & 0.63 \\  \hline 
\multirow{3}{*}{ SpectralClustering } & Activity Occurrence (Act) & 0.66 & 0.48 & 0.50 & 0.51 & 0.61 & 0.61 & 0.66 \\
& Global Features (Glo) & 0.66 & 0.55 & 0.64 & 0.62 & 0.63 & 0.72 & 0.66 \\
& Act - Glo & 0.62 & 0.46 & 0.50 & 0.50 & 0.57 & 0.61 & 0.62 \\ \hline 
\multirow{3}{*}{ Isolation Forest } & Activity Occurrence (Act) & 0.69 & 0.61 & 0.62 & 0.62 & 0.68 & 0.67 & 0.69 \\
& Global Features (Glo) & \textbf{0.76} & \textbf{0.68} & 0.71 & 0.68 & \textbf{0.74} & 0.75 & 0.76 \\
& Act - Glo & \textbf{0.76} & \textbf{0.68} & 0.71 & 0.68  & \textbf{0.74} & 0.75 & 0.76 \\ \hline 
\vspace{-1em}
\end{tabular}

}
\end{center}
\caption{Performance of the different methods implemented for the discovery of routine and non-routine days.}
\label{table:methodsPerformance}
\end{table*}

\begin{figure}[ht!]
    \centering 
    \includegraphics[width=\textwidth]{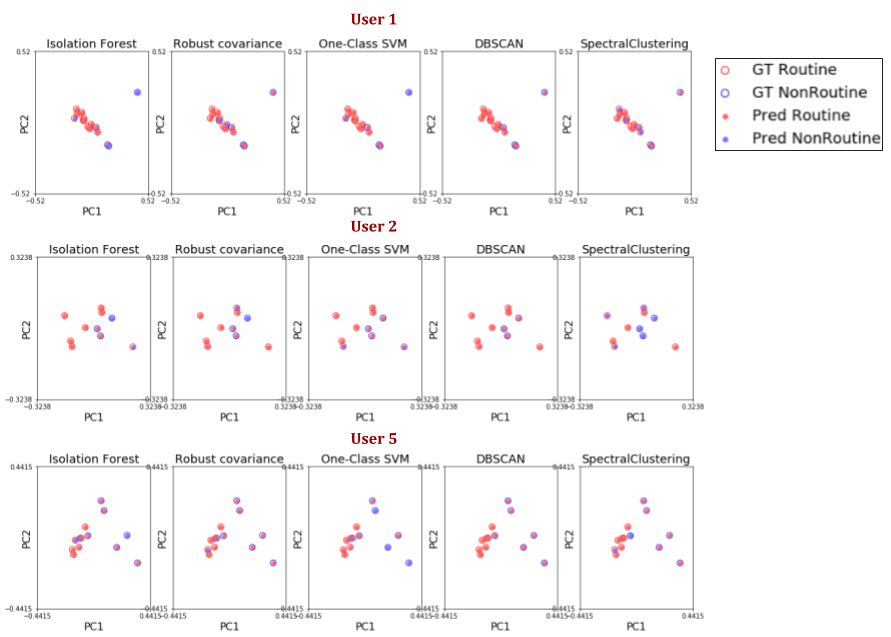}
    \caption{Visualization of the obtained classification results based on the analysis of the histogram of activities occurring throughout the day for User1, User2 and User5. We show the classification per user and per studied method. Each dot in the graph corresponds to one day recorded by the user. Each of the 4 subplots shows the classification into routine or non-routine by the baseline methods. The colour of the boundaries of the dots represents the given Ground-truth and the filling the classification label; Red routine and Blue non-routine.}
    \label{fig:plotresultsCAIP}
    \vspace{-1em}
\end{figure}

Moreover, in Fig. \ref{fig:plotresultsCAIP} we visualize the days as points in the feature space drawn by the first two principal components of the dataset. We can see the Ground-truth indicated with the boundaries of the circles and the prediction of the model, for both cases red corresponds to routine related days and blue to non-routine related. As it can be observed, our model is the one that obtains the best results. 

In Fig. \ref{fig:hist_act_Users} we can observe the occurrence of activities per day in the form of a histogram. This representation allows us to better infer and understand how routine (orange) and non-routine (blue) related days vary for the different camera wearers. From this representation we can confirm our initial assumptions: i) the set of activities performed as routine and non-routine related days differs per person, ii) a subset of activities is commonly shared when it comes to routine, such as `working', which is mostly described by a laptop/pc as central object in the scene, or `using mobile'. In contrast, some activities are specific per user: The routine of \textit{User 5} is characterized by `cooking', `reading', and `meeting'. In contrast, for \textit{User 2} `walking outdoor', `shopping', and `mobile' are the more representative activities.

\section{Conclusions}
\label{section:conclusions}
In this work, we propose a new unsupervised routine recognition model from egocentric photo-streams. To our knowledge, this is the first work on the field. Our proposed model achieves an average of 76\% Accuracy and 69\% Weighted F-Score for all the users. The results demonstrate the goodness of the proposed model and selected features. Moreover, it opens the door to the desired and personalized behavioural analysis.

The presented analysis can be improved in several directions as by augmenting the number of subjects and the amount of collected data. We believe this is a good starting point for this new field of unsupervised routine analysis from a first-person perspective. Moreover, and even though in this work we consider that there exists one routine per person, future lines will address the discovery of several routines. However, for that, it is needed a bigger amount of data. Currently, we are working on building a larger dataset with more users recording for larger periods of time. In future research, we will explore how the information of performed activities, and their temporal relation and context, can be integrated into the description of a day. The obtained information about routine related days and activities can have a direct positive impact on the lifestyle and health of the camera wearer. In addition, these findings provide additional information about how active a person is, which can be correlated to his or her emotional state. 

%
\section*{Acknowledgment}
This work was partially founded by projects TIN2015-66951-C2-1-R, 2017 SGR 1742, CERCA,  Nestore Validithi, 20141510 (La MaratoTV3) and CERCA Programme/Generalitat de Catalunya. P. Radeva is partially supported by ICREA Academia 2014. We acknowledge the support of NVIDIA Corporation with the donation of Titan Xp GPUs.

%
%
\bibliographystyle{splncs04}

\end{document}